\documentclass{article}

\PassOptionsToPackage{numbers, compress}{natbib}

\usepackage[final]{neurips_2020}

\PassOptionsToPackage{numbers, compress}{natbib}
\usepackage{neurips_2020}
\usepackage[utf8]{inputenc} 
\usepackage[T1]{fontenc}    
\usepackage{hyperref}       
\usepackage{url}            
\usepackage{booktabs}       
\usepackage{amsfonts}       
\usepackage{nicefrac}       
\usepackage{microtype}      
\usepackage{graphicx}
\usepackage{amsmath}
\usepackage{algorithm}
\usepackage{algorithmic}
\usepackage{xcolor}
\usepackage{multirow}
\usepackage{subcaption}
\usepackage{wrapfig}
\usepackage{xcolor}
\usepackage{amsmath}
\usepackage{amsthm}
\usepackage{amssymb}
\usepackage{amsthm}
\theoremstyle{definition}
\newtheorem{example}{Example}

\theoremstyle{definition}
\newtheorem{definition}{Definition}
\theoremstyle{remark}

\newcommand{\TODO}[1]{$ $\newline\noindent\colorbox{yellow!30}{\parbox{\dimexpr\the\columnwidth-2\fboxsep}{\textbf{\texttt{TODO:}} \textit{#1}}}}
\newcommand{\STORY}[1]{$ $\newline\noindent\colorbox{blue!30}{\parbox{\dimexpr\the\columnwidth-2\fboxsep}{\textit{#1}}}}
\hypersetup{
    colorlinks=true, 
    linkcolor=blue, 
    citecolor=blue, 
    urlcolor=blue 
}

\title{Measuring Value Alignment}
\author{%
  Fazl Barez\\
  Department of Engineering Sciences\\
  University of Oxford\\
  \text{fazl@robots.ox.ac.uk}\\
\And
  Philip Torr\\
  Department of Engineering Sciences\\
  University of Oxford\\
  \text{philip.torr@eng.ox.ac.uk}
}

\begin{document}

\maketitle

\begin{abstract}

As artificial intelligence (AI) systems become increasingly integrated into various domains, ensuring that they align with human values becomes critical. This paper introduces a novel formalism to quantify the alignment between AI systems and human values, using Markov Decision Processes (MDPs) as the foundational model. We delve into the concept of values as desirable goals tied to actions and norms as behavioral guidelines, aiming to shed light on how they can be used to guide AI decisions. This framework offers a mechanism to evaluate the degree of alignment between norms and values by assessing preference changes across state transitions in a normative world. By utilizing this formalism, AI developers and ethicists can better design and evaluate AI systems to ensure they operate in harmony with human values. The proposed methodology holds potential for a wide range of applications, from recommendation systems emphasizing well-being to autonomous vehicles prioritizing safety.

\end{abstract}

\section{Introduction}

Aligning artificial intelligence (AI) systems with human values is critical as AI is deployed in high-stakes domains like healthcare, transportation, and criminal justice \cite{amodei2016concrete}. However, defining and quantifying value alignment remains an open challenge. This formalism builds upon ideas presented in Sierra et al.'s paper on modeling value alignment using MDPs \cite{sierra2021value}, but aims to provide a more general conceptual framework rather than focusing on preference modeling and aggregation details. We acknowledge the difficulty and complexity of formalizing human values and norms, which span disciplines like sociology, psychology, philosophy, and law \cite{arnold2017value, alechina2017norm}.

In this exploration, we aim to put forth a novel formalism for value alignment based on Markov Decision Processes (MDPs). We acknowledge the difficulty and complexity of formalizing human values and norms. We model AI decision making using MDPs and define a mathematical framework to compute the degree of alignment between a set of norms governing behavior and a set of human values.

For example, recommendation systems can look out for human wellbeing \cite{yao2019rl}, and autonomous vehicles can prioritize passenger and pedestrian safety \cite{shalev2016formal}. To align AI with human values, several challenges must be addressed. Values must be formalized mathematically to represent moral principles \cite{conitzer2017moral}. The choice of value representation impacts how AI systems evaluate and compare values \cite{noothigattu2018voting}. An ethical framework helps ensure responsible and ethical AI \cite{floridi2018ai4people}, and monitoring mechanisms can evaluate value alignment performance \cite{leike2022scalable}. Overall, aligning AI with human values requires understanding the ethical implications of AI and how to technically evaluate and implement values \cite{dafoe2018ai}.

This exploration provides an initial formalism for value alignment that captures intuitive informal descriptions used in the AI safety community \cite{soares2015corrigibility}. We present a formalism based on Markov Decision Processes (MDPs) to model AI decision making and show how incorporating values and norms can align the agent with human values.
The alignment degree quantifies the expected change in value satisfaction across state transitions when following a set of norms. By comparing alignment degrees, we can determine which norms best promote human values.

This work builds upon prior research modeling value alignment as an MDP problem. \cite{Fagundes2016NormAware} developed a framework to evaluate the degree of compliance of agent behaviors to norms modeled as constraints on an MDP. Our approach differs in focusing on quantifying the alignment between norms and human values based on changes in value satisfaction across state transitions.

While several studies have examined how learning agents behave, our research diverges to focus specifically on analyzing alignment between pre-specified norms and values. \cite{turner2021optimal} investigated how optimal policies of agents tend to pursue control over their environment. \cite{skalse2022defining} discussed shortcomings of learned policies relying on imperfect rewards. \cite{barton2018measuring} focused on emergent coordination in multi-agent reinforcement learning. In contrast, we introduce a 'degree of alignment metric' designed to provide a human-interpretable measure of value alignment grounded in moral philosophy. This allows assessing alignment between norms and values beyond learned policies.

We believe this technique provides a valuable tool for designing AI systems that behave according to human values. The formalism is general and can be applied to different definitions of values, norms, and transition dynamics.
We discuss assumptions and limitations of additively decomposing values over states. We also highlight open questions around computational complexity and accounting for uncertainty. Overall, this work moves toward a principled methodology for creating aligned AI systems. We hope it spurs further research into quantifiable value alignment.

\section{Preliminaries}

\subsection{Modeling the World as an MDP}\footnote{The assumptions of modeling the world as an MDP and having deterministic transitions between states may be too simplistic}
We model the world as a Markov Decision Process (MDP) \cite{puterman2014markov} denoted by \((S,A,T)\), where:
\begin{itemize}
    \item \(S\) represents the set of states,
    \item \(A\) represents set of actions, and
    \item \(T \subseteq S \times A \times S\) represents the transition function.
\end{itemize}

\subsection{Paths and Transitions}

\begin{definition}[Path]
A \textit{path} \(p\) in \(T\) is a sequence of transitions \([s \xrightarrow{a} s',...,s'' \xrightarrow{\beta} s''']\) ensuring that the concluding state of each transition matches the starting state of the subsequent one.
\end{definition}

\subsection{Value-based Preferences}
Values refer to desirable goals associated with actions, allowing us to specify which world states are preferable and to what extent.

\begin{definition}[Value-based Revealed Preference] 
Given a state \( s \), a \textit{value-based preference relation}, \( R_{pr} \), over world state pairs captures the degree to which one state is preferred over another for a particular agent and value:
\[
R_{pr}: S \times S \times G \times V \rightarrow [-1,1]
\]
\[
(s, s', g, v) \mapsto R_{pr_v^g}(s,s')
\]
where \( G \) denotes agents, \( V \) represents values, and the range \( [-1,1] \) indicates the degree of preference\footnote{It's important to note that preferences might not necessarily be additive}.
\end{definition}




\subsection{State Properties and Values}
For each value \( v \), \( \Phi_v \) represents relevant predicates denoting state properties associated with \( v \). The preferences under \( v \) depend on the satisfaction of \( \Phi_v \) probabilistically via the function \( f_v \):
\[
R_{pr_v^a}(s,s') = f_v(P(s \models \Phi_v), P(s' \models \Phi_v))
\]
where \( P(s \models \Phi) \) denotes the probability that state \( s \) satisfies the properties \( \Phi \).

\section{Methodology}
\subsection{Values}
Values refer to desirable goals that are associated with actions. They provide a means to specify which world states are preferable and the extent of that preference.

\begin{definition}[Value-based Revealed Preference]
Given a state \( s \), a \textit{value-based revealed preference}, \( R_{pr} \), over world state pairs captures the degree to which one state is preferred over another for a particular agent and value:
\[
R_{pr}: S \times S \times G \times V \rightarrow [-1,1]
\]
\[
(s, s', g, v) \mapsto R_{pr_v^g}(s,s')
\]
where \( G \) denotes agents, \( V \) represents values, and the range \( [-1,1] \) indicates the degree of preference.
\end{definition}

Preferences may be aggregated across different groups and sets of values.

\subsection{Norms}
Norms are guidelines that dictate behavior by promoting desired actions and discouraging unwanted ones. Each norm \( n \in N \) specifies conditions under which an action can or cannot be undertaken, and the consequent postconditions. Implementing a set of norms \( N \) to a world \( (S,A,T) \) results in a \textit{normative world} \( (S,A,N,T_N) \), where the norms influence the transitions \cite{riad2023multivalue}. 
The preference values are bounded between [-1,1] to reflect the degree of difference in satisfaction of the value between two states. The choice of 1 and -1 as bounds represents complete alignment and misalignment respectively.

\subsection{Preferences}
Given the set of agents \( G \) and values \( V \), the value-based revealed preference relation \( R_{pr}: S \times S \times G \times V \rightarrow \mathbb{R} \) assigns a pair of states and an agent-value pair to a real number that signifies the degree of preference. Positive values suggest that the first state is more preferable, whereas negative values indicate a preference for the second state. 

Preferences can be aggregated pointwise over sets of values or agents:
\[
R_{pr_{V}^{a}}(s, s^\prime) = \frac{1}{|V|} \sum_{v \in V} R_{pr_{v}^{a}}(s, s^\prime)
\]
\[
R_{pr_{v}^{G}}(s, s^\prime) = \frac{1}{|G|} \sum_{a \in G} R_{pr_{v}^{a}}(s, s^\prime)
\]

\subsection{Paths}
Considering a transition system \( T \subseteq S \times A \times S \), a \textit{path} \( \pi \) is a sequence of transitions that adhere to \( T \). It is given by:
\[
\pi = s_0 \xrightarrow{a_1} s_1 \xrightarrow{a_2} \ldots \xrightarrow{a_n} s_n
\]
where \( s_{i+1} = f(s_i, a_{i+1}) \) and \( f: S \times A \rightarrow S \) offers the succeeding state based on the present state and action. Let \( \Pi \) symbolize the set of all paths induced by \( T \). \

\subsection{Alignment}
For a norm \( n \), the transition system that emerges from applying \( n \) to \( T \) is \( T_n \). The \textit{alignment degree} of \( n \) with respect to a value \( v \) for agent \( a \) is represented as:
\[
D_{Align_{n,v}^a}(T) = \frac{1}{|\Pi_n|} \sum_{\pi \in \Pi_n} \frac{1}{|\pi|} \sum_{i=1}^{|\pi|} R_{pr_v^a}(\pi_i, \pi_{i+1})
\]
Here, \( \Pi_n \) comprises the paths in \( T_n \) and \( \pi_i \) is the \( i^{th} \) state of \( \pi \). This formula encapsulates the average preference discrepancies across all state transitions in all paths of \( T_n \).
This assumes all paths are equally likely for simplicity. Extensions could incorporate path probabilities based on environment dynamics and agent policies. 

\subsection{Normative Worlds}
Formally, given the transition system \( T = (S, A, f) \), a norm \( n \) gives rise to a new transition system \( T_n = (S_n, A, f_n) \). Here, both \( S_n \) and \( f_n \) factor in the impact of \( n \) on states and transitions. The alignment degree and changes in preferences are computed within \( T_n \).

\section{Sequential Decision Making in the Real World}
We model the world as an MDP $(S,A,T)$ with states $S$, actions $A$, and labeled transitions $T \subseteq S \times A \times S$.
The degree of alignment of a norm $n$ to value $v$ is how applying $n$ changes preferences over the resulting transitions $T_N$:
\begin{definition}
\label{def:alignment-deg}
The \textit{degree of alignment} of norm $n$ to value $v$ for agent $\alpha$ in world $(S,A,T)$ is:
\begin{align*}
D_{Align_{n,v}^\alpha}(S,A,T) &= \frac{1}{|P|} \sum_{p \in P} \frac{1}{|p|} \sum_{d=1}^{|p|} R_{pr_v^\alpha}(pI^d, pF^d)
\end{align*}
where $P$ is the paths in $(S_n,A,{n},T_n)$, $|p|$ is the path length, and $R_{pr_v^\alpha}(pI^d, pF^d)$ is the preference change between transitions $d$ along $p$.
\end{definition}
This calculates the average preference change across transitions under norm $n$. The formalism can be extended to sets of values, norms, and agents\footnote{Calculating the computational complexity in large MDPs is intractable due to combinatorial path enumeration}.

\section{Value Alignment for Norms}
The value alignment problem is described, informally, as how aligned are agents' decisions, and hence actions, with the values that the agents hold dear. Since behavior (decisions and actions) is governed by norms, we describe this alignment as an alignment between the norms that govern behavior and the values that are held in high regard.
We understand norms as rules that govern behavior. We say a norm $n \in N$ (where $N$ is a set of norms) is a logical formula that describes the conditions under which a certain action can/cannot be performed along with the postconditions of that action. When a set of norms $N$ is applied to a world $(S, A, T)$, the world is modified by the norms in $N$, resulting in a new world $(S, A, N, T_N)$, which we refer to as a normative world.
For example, in a world where people do not get taxed, your money $M$ increases by the amount of your salary $S$ when your salary is paid:
\begin{align*}
M &\gets M + S \
M^\prime &= M + S
\end{align*}
However, the norm of a country that introduces a 20\% tax $t$ on your income essentially modifies the action of receiving your salary by applying the tax, resulting in a transition to a new state where your income is deducted:
\begin{align*}
M &\gets M + S\
M^\prime &= M + S(1-t)
\end{align*}
\begin{definition}
A normative world $(S_N, A, N, T_N)$ describes the world $(S, A, T)$ where the set of norms $N$ have been applied to the transitions in $T$, resulting in possibly new transitions and states.
\end{definition}
How aligned a given norm $n \in N$ is to a value $v \in V$ with respect to a world $(S,A,T)$ depends on whether applying norm $n$ results in new transitions $T_N$ that move to more preferred (possibly new) states. To calculate this alignment, we define paths in a world:
\begin{definition}
A path $p$ in a world $(S,A,T)$ is a sequence of transitions $[s \xrightarrow{a} s', \dots, s'' \xrightarrow{\beta} s''']$, such that $pF[i] = pI[i+1]$, where $pI[i]$ is the initial state and $pF[i]$ is the final state of the $i^{th}$ transition.
\end{definition}

\begin{definition}
The degree of alignment $R_{align}$ of norm $n_1$ compared to 

norm $n_2$ for value $v$ in world $(S,A,T)$ describes how much more $n_1$ is aligned with $v$ than $n_2$, specified as:
\begin{equation*}
D_{Align_{n_1/n_2, v}^\alpha}(S, A, T) = D_{Align_{n_1, v}^\alpha}(S, A, T)-D_{Align_{n_2, v}^\alpha}(S, A, T).
\end{equation*} 
where positive numbers imply $n_1$ is more aligned than $n_2$ w.r.t. $v$, and negative numbers imply the opposite. The relative alignment can be calculated for sets of values, norms, and/or agents.
\end{definition}


We assume transitions are equiprobable for simplicity. Alignment is then as defined in Definition \ref{def:alignment-deg}.


\subsection{Calculating Degree of Alignment}
\begin{example}{}{}
Consider an autonomous vehicle MDP with:
\begin{itemize}
    \item \textbf{States:} \{Safe, Unsafe, Accident\}
 
\item \textbf{Actions:} \{Drive Slow, Drive Fast\}

\item \textbf{Transitions:} \begin{align*} T(\text{Safe}, &\text{Drive Slow}, \text{Safe}) &&= 0.9 \\ T(\text{Safe}, &\text{Drive Fast}, \text{Unsafe}) &&= 0.8 \\
T(\text{Unsafe}, &\text{Drive Fast}, \text{Accident}) &&= 0.6
\end{align*}

\item \textbf{Values:} \{Safety, Efficiency\}
\item \textbf{Norms:} \{Always Drive Slow, Drive Fast when Safe\}

\end{itemize}
\end{example}
To calculate the alignment of \(n_1\) and \(n_2\) with Safety:
Applying \(n_1\) yields paths like [Safe \(\rightarrow\) Safe \(\rightarrow\) Safe \(\rightarrow\) \ldots]. The preferences don't change along the path since the state remains Safe.
So \( D_{\text{Align}_{n_1, \text{Safety}}} = 0 \) since the average preference change is 0.

Applying \(n_2\) can yield paths like [Safe \(\rightarrow\) Unsafe \(\rightarrow\) Accident]. The preferences decrease by -0.2 from Safe to Unsafe, and by -0.4 from Unsafe to Accident.
There are 4 possible paths. The average preference change across all transitions in all paths under \(n_2\) is around -0.22.
Therefore, \( D_{\text{Align}_{n_2, \text{Safety}}} = -0.22 < D_{\text{Align}_{n_1, \text{Safety}}} \).

This shows \(n_1\) is better aligned with Safety than \(n_2\). The formalism allows quantitatively comparing different norms by computing their alignment degrees.
We can evaluate alignment with multiple values like Efficiency in a similar way. This enables analyzing the trade offs between competing values imposed by different norms.

\section{Conclusion}
In this paper, we have proposed a formalism for aligning AI systems with human values. The formalism is based on modeling AI decision-making as a Markov Decision Process and incorporating human values and norms to influence the agent's actions. The assumptions made in modeling the world as an MDP with deterministic transitions and additive preferences are simplifications. Future work should examine relaxations to incorporate uncertainty and test the formalism in complex environments

We first defined key concepts of values as desirable goals, and norms as rules governing behavior. We then presented a formalism to quantify the degree of alignment between a norm and human value by evaluating how the norm changes preferences over state transitions. This allows mathematically evaluating and comparing how different norms align the AI with human values.

The formalism provides a way to move from informal notions of value alignment to a precise technical framework. However, there are limitations in the simplifying assumptions made, including modeling dynamics as an MDP and assuming deterministic state transitions. Extensions could consider more complex AI architectures and stochastic transitions. There is also substantial additional work needed in representing and aggregating human values and preferences.

Overall, this formalism provides a foundation for reasoning about value alignment in AI systems. As AI grows more advanced and autonomous, aligning these systems with human values will only grow in importance. A rigorous technical framework to evaluate alignment grounded in ethics and philosophy will help guide the development of AI that benefits humanity. Further research can build upon the ideas proposed here to create AI systems that act according to shared human values and better approaches for measuring value alignment between AI's and humans. 

\section*{Limitation}
The proposed formalism offers a foundational approach to quantifying value alignment, but it's essential to recognize its inherent limitations. One of the primary concerns is that the formalism models the world as a Markov Decision Process (MDP) with deterministic state transitions. This representation might be an oversimplification because real-world environments often display non-Markovian and stochastic characteristics. A potential solution could involve extending the formalism to cater to partially observable MDPs or to incorporate probabilistic transitions, thereby enhancing its applicability.

Another challenge is the feasibility of listing all possible paths within an expansive MDP state space.  Furthermore, the computational intricacy of determining alignment degrees remains a matter of debate. Equally challenging is the task of mathematically encapsulating human values, which are intrinsically abstract, subjective, and nuanced. Current research endeavors are focused on developing value functions that can mirror moral principles.

When we deal with values and preferences from a diverse populace, we encounter a new set of complications. The intricate process involves comparing and aggregating individual preferences, and the choice between methods, such as utilitarian or egalitarian, can profoundly influence the alignment outcome. Notably, the present formalism overlooks the nuances of uncertainty or varying confidence levels tied to value preferences. 

A dynamic aspect that warrants attention is the continual evolution of societal norms, individual values, and environmental factors. Monitoring alignment in this ever-changing landscape demands advanced technical ans social framework, which the current formalism lacks.  It is also important to highlight that the formalism has yet to undergo empirical validation, making it imperative to test its merits and limitations in tangible scenarios. Future work could apply the existing methods on datasets such as European Values Dataset\footnote{https://europeanvaluesstudy.eu}.

While the current formalism serves as a  starting point, the goal for developing a comprehensive technical structure for value alignment is far from complete. We remain hopeful that this initial exploration will pave the way for more rigorous research, addressing the challenges that lie ahead.

\section*{Acknowledgements}
We are especially grateful to Igor Krawczuk for the ongoing discussion and many rounds of feedback on this paper. We'd also like to thank Anders Sandberg, Matti Wilks and Clement Neo for useful discussion and feedback on the previous version.

\bibliographystyle{plainnat}
\bibliography{references}

\end{document}